\documentclass[letterpaper]{article}
\usepackage{imav}
\usepackage{makecell, tabularx}
\usepackage{tabularray}
\usepackage{float}
\restylefloat{table}
\usepackage{times}
\usepackage{graphicx}
\usepackage[dvipsnames]{xcolor}
\usepackage{array}
\usepackage{booktabs}
\usepackage{multirow}
\usepackage{authblk}
\usepackage[dvipsnames]{xcolor}
\usepackage{tikz}
\usepackage{hyperref}
\hypersetup{
    colorlinks=true,
    linkcolor=blue,
    filecolor=magenta,      
    urlcolor=cyan,
    pdftitle={Overleaf Example},
    pdfpagemode=FullScreen,
    }

\newcolumntype{P}[1]{>{\centering\arraybackslash}p{#1}}
\newcolumntype{C}{>{\centering\arraybackslash}X}

\usepackage{wasysym}

\title{Exploring the potential of collaborative UAV 3D mapping in Kenyan savanna for wildlife research}

\author{ Vandita Shukla\textsuperscript{1,2}\thanks{vshukla@fbk.eu},  Luca Morelli\textsuperscript{1,3} \thanks{lmorelli@fbk.eu}, Pawel Trybala\textsuperscript{1}\thanks{ptrybala@fbk.eu}, Fabio Remondino\textsuperscript{1}\thanks{remondino@fbk.eu}, Wentian Gan\textsuperscript{4}\thanks{gwt2019@whu.edu.cn}, Yifei Yu\textsuperscript{4}\thanks{yfyu2020@whu.edu.cn} and Xin Wang\textsuperscript{4}\thanks{xwang@sgg.whu.edu.cn}}

\affil{\textsuperscript{1}3D Optical Metrology Unit (3DOM), Bruno Kessler Foundation (FBK), Trento, Italy \\ \textsuperscript{2} Computer Vision and Machine Learning Systems Group, Institute for Geoinformatics, University of Muenster, Germany \\ \textsuperscript{3}Dept. of Civil, Environmental and Mechanical Engineering, University of Trento, Italy \\  \textsuperscript{4} School of Geodesy and Geomatics, Wuhan University, People’s Republic of China }

\begin{document}

\maketitle
\thispagestyle{empty} 
\begin{abstract}
UAV-based biodiversity conservation applications have exhibited many data acquisition advantages for researchers. UAV platforms with embedded data processing hardware can support conservation challenges through 3D habitat mapping, surveillance and monitoring solutions. High-quality real-time scene reconstruction as well as real-time UAV localization can optimize the exploration vs exploitation balance of single or collaborative mission. In this work, we explore the potential of two collaborative frameworks - Visual Simultaneous Localization and Mapping (V-SLAM) and Structure-from-Motion (SfM) for 3D mapping purposes and compare results with standard offline approaches.
\end{abstract}

\section{Introduction} \label{section:introduction}
Unmanned Aerial Vehicles (UAVs) have become an essential tool for supporting conservation challenges to collect data for 3D mapping, surveillance and monitoring \cite{tuia_perspectives_2022, wirsing_2022, shapiro_conservation_2020, shukla_towards_2024}. UAVs, however, have generally limited flying time due to battery power, requiring solutions that increase the efficiency of data collection in the limited airborne time. A possible solution is to use multiple UAVs (or agents/swarm) which can collaborate in the data collection to support studies on wildlife populations and habitat \cite{barbeau_research_2022} and overcome the challenges induced by the use of a single platform through collaborative missions \cite{10318509}. Collaborative mapping has shown its potential to study wide-ranging animal species that require larger area coverage or locating them in complex environments that may require a longer survey time \cite{huang_multi-uav_2022}\cite{kabir_wildlife_2021}. Moreover, different agents in the mission can leverage different hardware and sensor strengths for fulfilling multiple goal targets \cite{tong_multi-uav_2023}\cite{chen_overview_2023} \cite{10.1145/3150165.3150166}.  Within the context of wildlife monitoring, a UAV-based collaborative approach has multiple benefits, including:
\begin{itemize}
    \item extension of mission time of an on-going pursuit (e.g. real-time poacher or animal tracking);
    \item 3D geometric information recovery (e.g. 3D pose estimation from stereo vision from two synchronised UAVs);  
    \item coverage of wider target area (e.g. for tracking multiple herds or speeding up mapping operations).
\end{itemize}

\par UAV surveys for wildlife studies have multifaceted goals \cite{quantifying}, such as mapping the terrain while keeping a lookout for dynamic targets such as animal herds or poachers. Missions need frequent path re-planning due to changes in the objective priority, which can switch from mapping the scene to tracking targets. Therefore, upon locating an animal group of interest, diverting from original waypoint-based mapping path and navigating to unknown terrains is often required when the animals move. In such a scenario where one UAV has to divert to follow animals of interest, continuation of original survey can be achieved through collaborative mapping i.e. a follow-up from one or more UAVs. High quality real-time scene reconstruction and camera trajectory feedback can optimize the \textit{exploration} vs \textit{exploitation} balance of such a collaborative mission through enhanced situational awareness for each agent. Exploration and exploitation are the key phases of collaborative missions in which the information gained by one or more UAVs (\textit{exploration}) is used for path planning optimization (\textit{exploitation}) to achieve data acquisition with the least redundancy and best resource utilization \cite{barbeau_research_2022}. 
\par
 Visual Simultaneous Localization and Mapping (V-SLAM) is a key technique in UAV-based 3D mapping. It allows an agent to localize itself within a 3D space in real-time while reconstructing the map (environment) through various sensors \cite{wang_uav-based_2024}. SLAM saw increase in popularity for visual sensor based mapping in 2004 with the showcase of MonoSLAM \cite{davison_monoslam_2007}, a first complete real-time visual-SLAM system. A SLAM system consists of the following parts: front-end comprising sensor and odometry components; and back-end for loop optimization and mapping. Front-end acquires data from single or multiple agents\cite{ahmed_active_2023} and, in case of V-SLAM, covers image processing (feature extraction, matching and tracking, and pose estimation on existing map), while back-end increases and optimizes the map. Multi-agent visual SLAM architectures are usually centralized: UAVs or unmanned ground vehicles (UGVs) acting as front-end agents focus on their real-time state estimation, and a server backend that performs aggregate map generation and loop closure. Collaborative V-SLAM's performance deteriorates in complex outdoor and dynamic environments, especially if RGB cameras are the unique odometry sensors. While exploring larger areas in savanna with multiple UAV agents, repetitive features and less flight time can lead to absence of loop closures causing the drift of the trajectory to increase rapidly uncorrected. Systems equipped with additional types of sensors, such as IMU, can be more reliable for mapping outdoor complex environments \cite{wang_uav-based_2024, cao_scaling_2024, chen_overview_2023, tong_multi-uav_2023, danping_collaborative_2019}. 
 
 \par Starting from image sequences, Structure-from-Motion (SfM) estimates three-dimensional structures, usually a sparse map of 3D tie points, and image poses \cite{jiang_efficient_2020}. When it comes to high-fidelity mapping with RGB data, SfM can be more accurate than V-SLAM, e.g. since SLAM forces a limit on the number of tie points and/or uses lower quality local features to improve computational time. However, traditional SfM is generally performed offline  \cite{reja_computer_2022}, in contrast to the real-time-oriented V-SLAM. Recent advancements propose real-time Structure-from-Motion, also called On-the-Fly (OtF) SfM \cite{zhan_--fly_2024}. OtF-SfM has demonstrated the capability to photogrammetrically process image streams acquired by multiple collaborative agents without the requirement of spatio-temporally ordered input, normally a major pre-requisite for onboard V-SLAM \cite{9812319}. 
 The performance of collaborative V-SLAM depends on individual UAV mapping performance, since the SLAM algorithm runs onboard each UAV independently, whereas optimization such as loop-closures and map-aggregation happen on a centralized back-end \cite{cao_scaling_2024, danping_collaborative_2019}. On the other hand, collaborative SfM could reduce the dependence of collaborative mapping on individual agent mapping accuracy or on inter-agent communication for global optimization \cite{cao_scaling_2024}. It can handle asynchronous image input from multiple UAVs and prior reconstructions are seamlessly merged on the server into a complete model whenever images can be registered, which motivates evaluation of its performance in challenging savanna environment.
\par In this paper, we aim to study the potential and limitations of UAV-based collaborative mapping in the context of wildlife conservation using only visual data. A real-time SfM approach (hereafter referred to as OtF-SfM \footnote{\url{https://yifeiyu225.github.io/on-the-flySfMv2.github.io/}}) and a collaborative V-SLAM software (hereafter referred to as CCM-SLAM \cite{schmuck_ccm-slam_2019}), have been compared to traditional offline SfM processing. In addition, the influence of learning-based tie points extracted with convolutional neural networks (CNNs)\cite{morelli2024deep} trained for difficult scenarios is tested evaluating the final accuracy of flight trajectories.

\section{Data} \label{section:data}
The data used for the study was collected in July 2023 at the Ol Pejeta Conservancy, located in Laikipia County (Kenya), as part of the preliminary data collection mission of the WildDrone\footnote{\url{https://wilddrone.eu}} EU project.

\begin{table}
  [H] 
  \setlength{\extrarowheight}{5pt}
  \begin{tabular}
      {P{1cm}P{3cm}P{3cm}}
      \hline \textbf{Dataset ID} & \textbf{Sample Image} & \textbf{Trajectories} \\ \hline\\
      \textbf{1} 
      & 
      \parbox[c]{3cm}{\centering
      \includegraphics[scale=0.045]{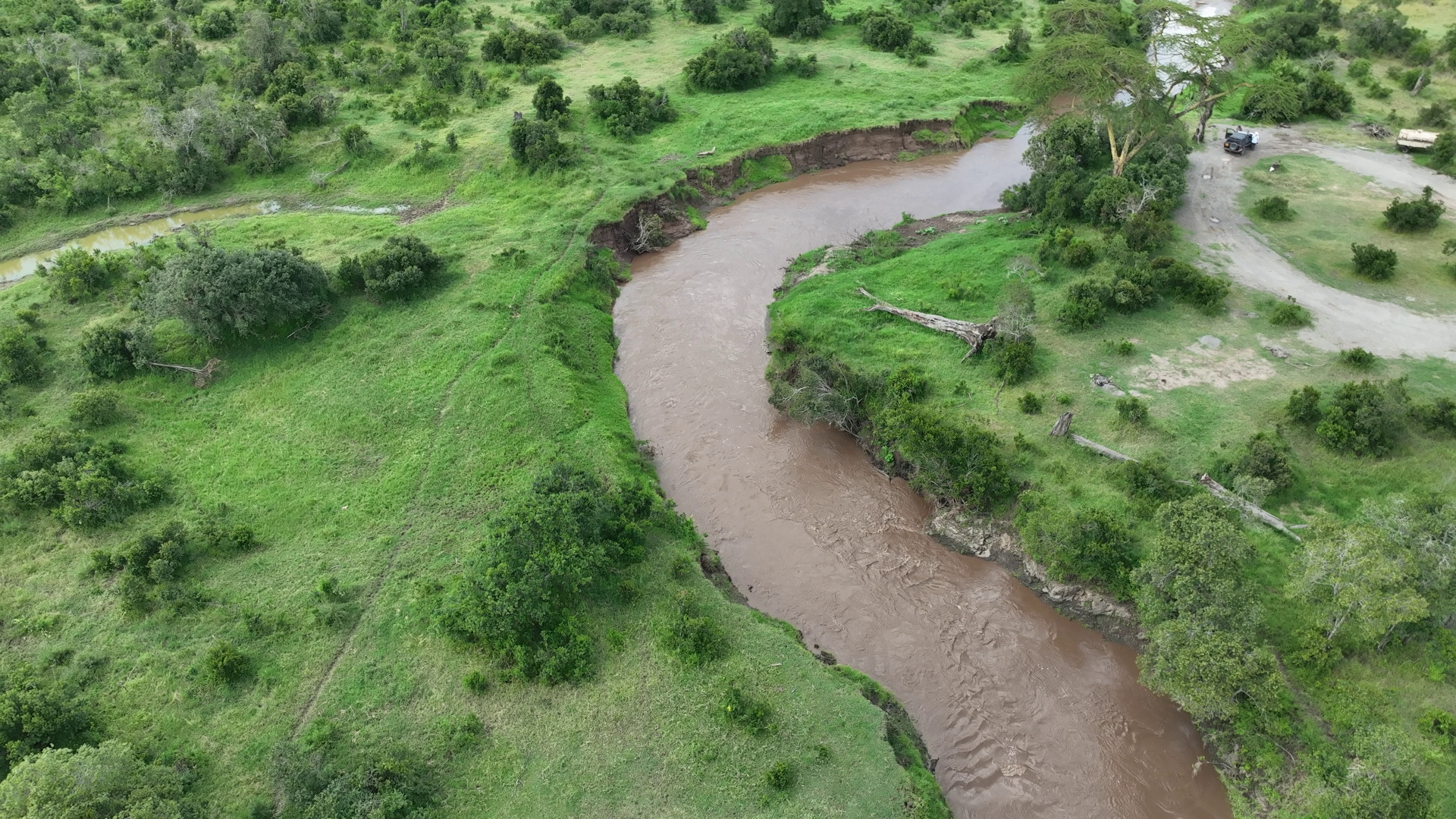}\\
      \includegraphics[scale=0.045]{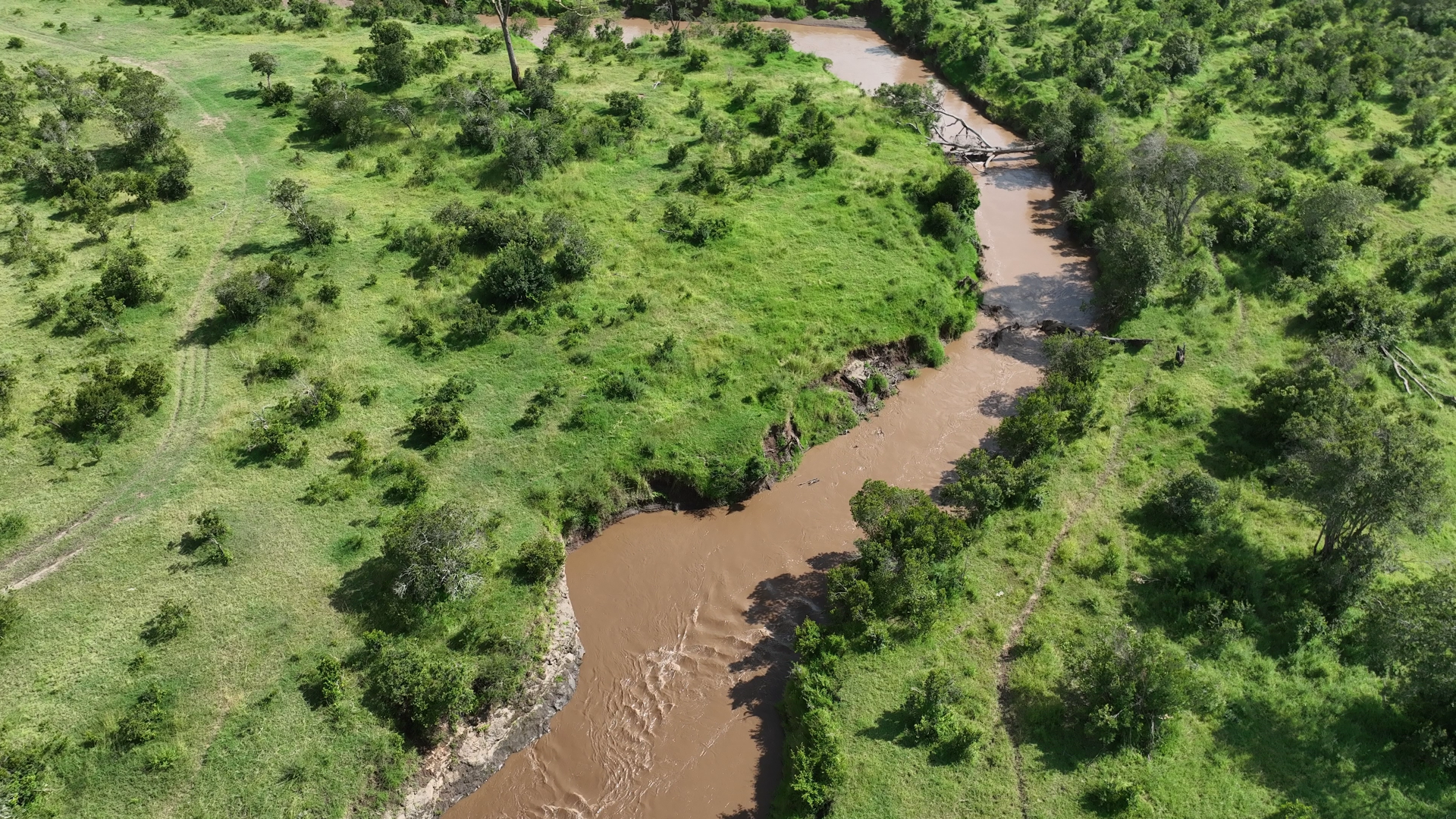}}
      & \parbox{3.5cm}{\centering
      \includegraphics[scale=0.139]{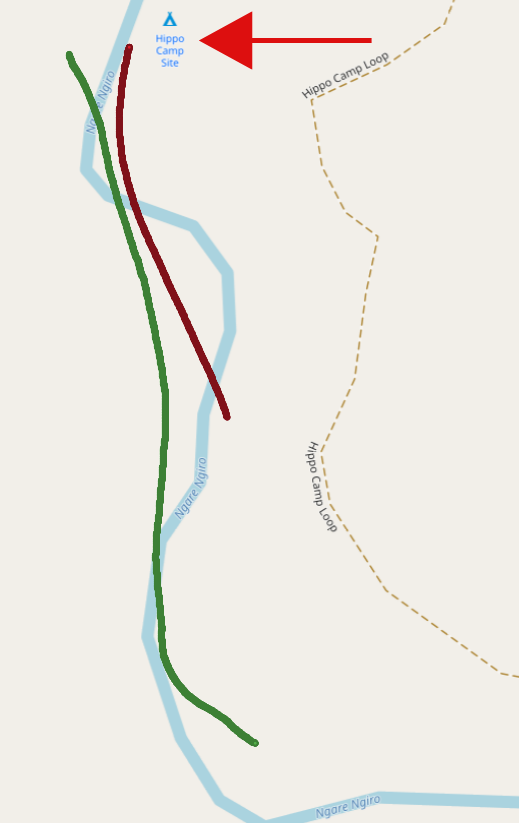}}\\
      \\\hline\\ 
      \textbf{2} & \parbox[c]{3cm}{
      \includegraphics[scale=0.045]{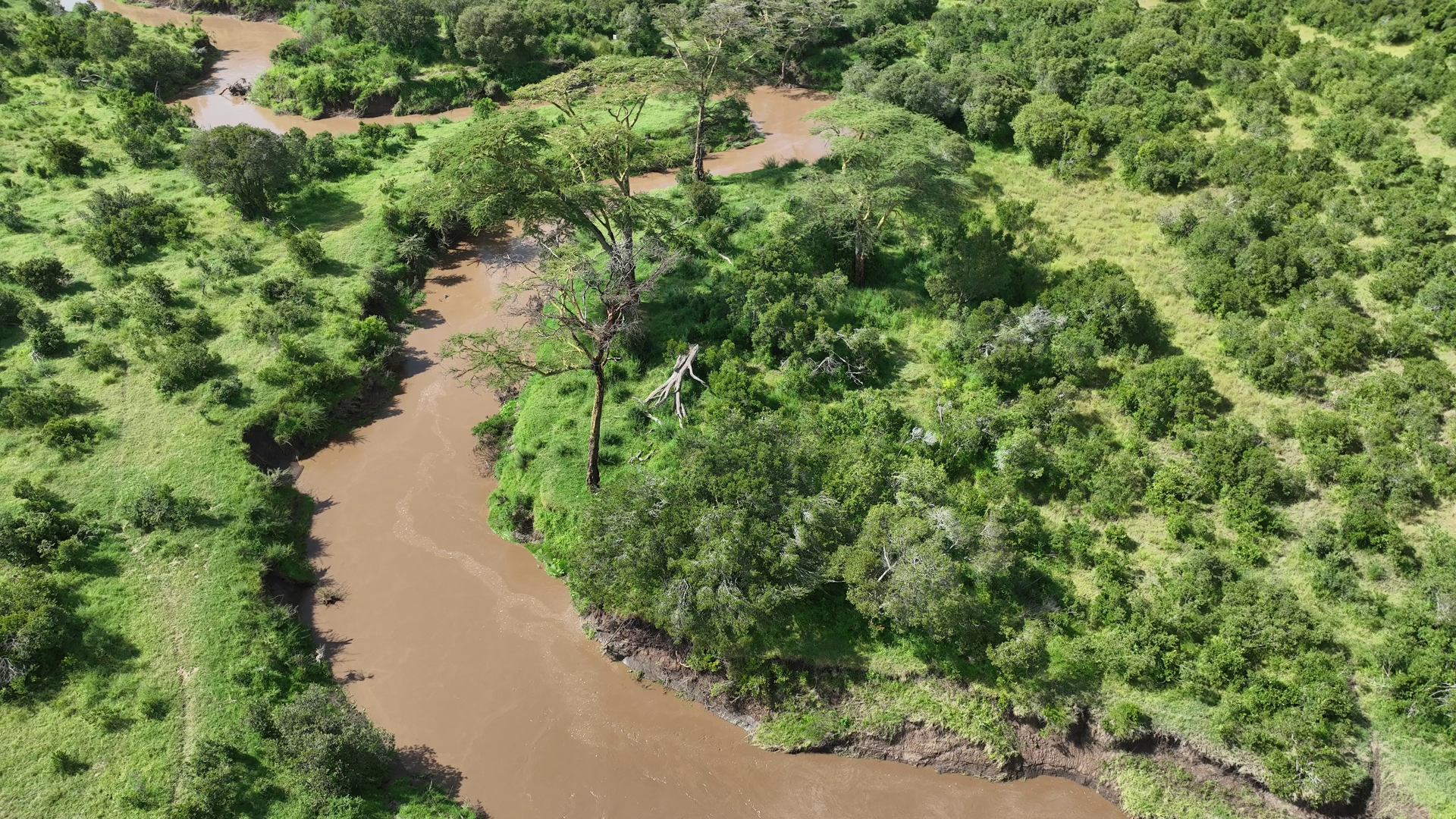}\\
      \includegraphics[scale=0.045]{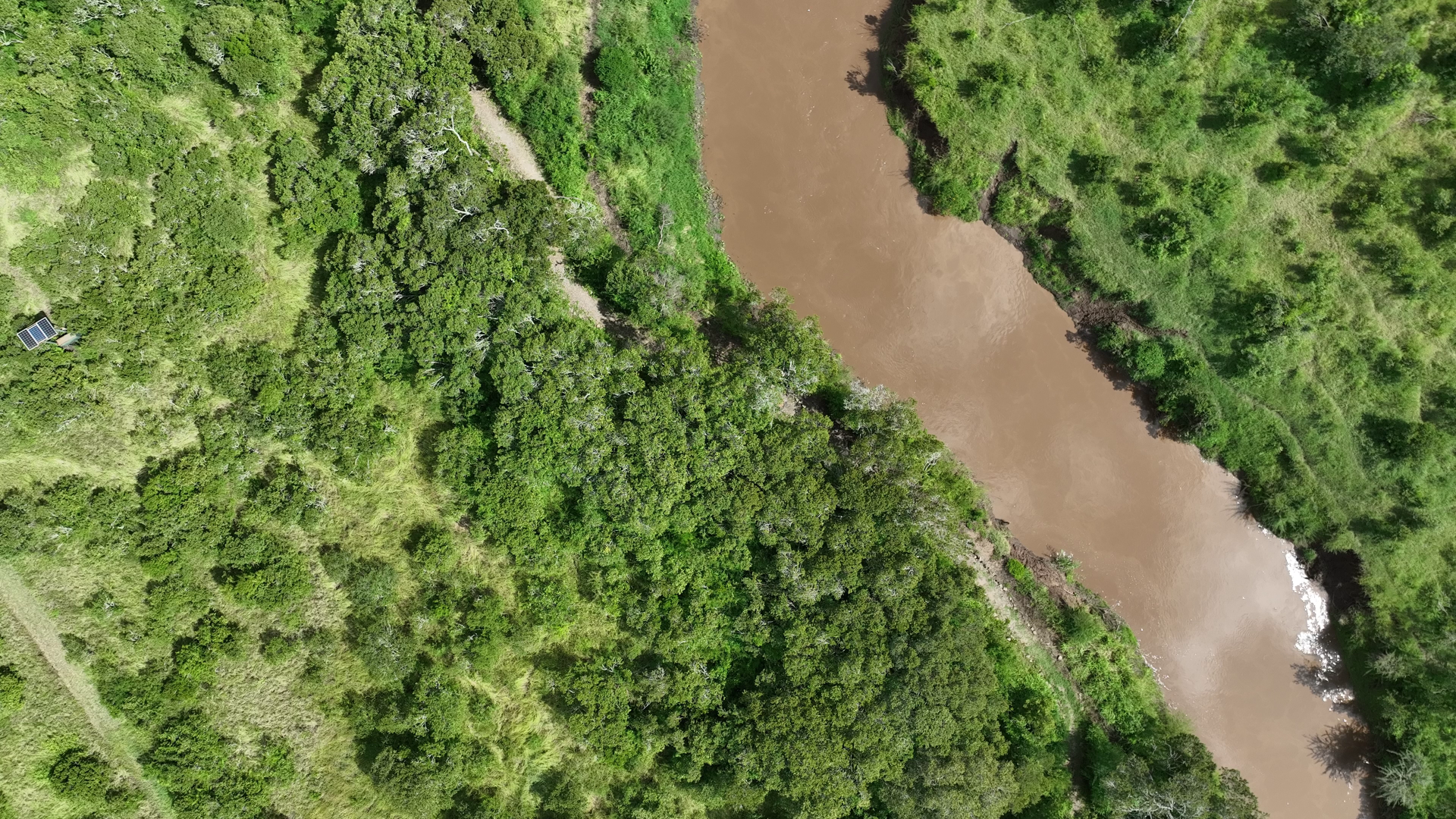}\\
      \includegraphics[scale=0.045]{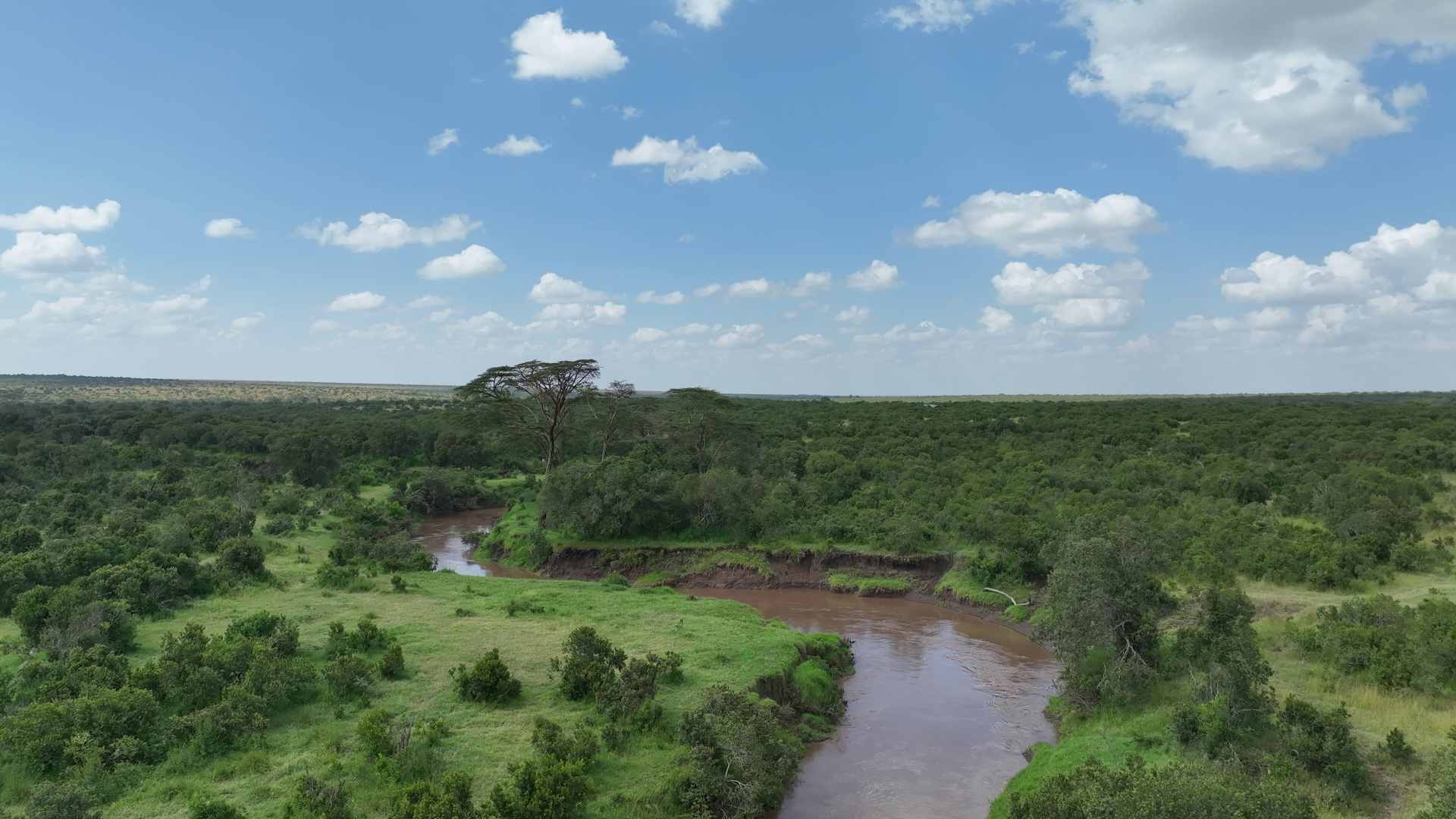}}
      & \parbox{3.5cm}{
      \centering\includegraphics[scale=0.135]{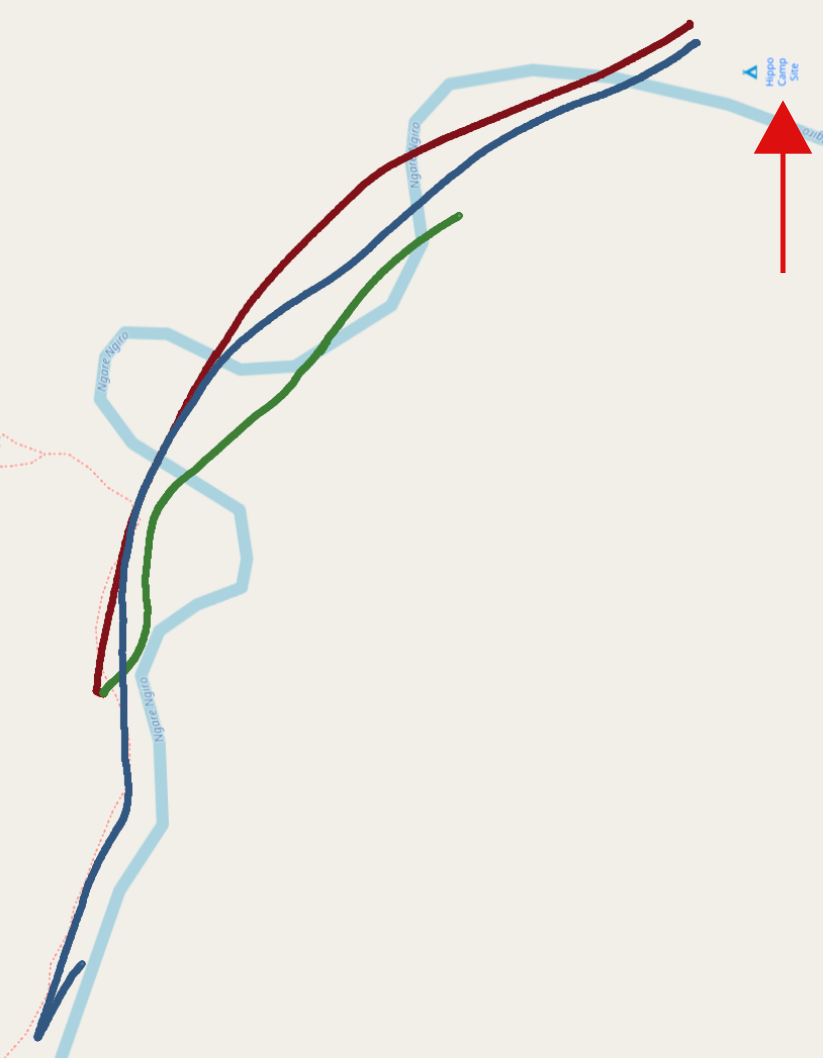}}\\
      \\\hline
      \multicolumn{3}{|c|}{\textbf{\textcolor{BrickRed}{red}} - flight 1 / \textbf{\textcolor{OliveGreen}{green}} - flight 2 / \textbf{\textcolor{BlueViolet}{blue}} - flight 3} \\ \hline\\
  \end{tabular}
  \caption{Used datasets with image samples and GNSS trajectories (L1 GPS positions stored by the onboard receiver) - different colors represent different agents. The red arrow points at the "Hippo Camp" on the map, a location later used to describe the flight directions.} 
  \label{table:used_data}
\end{table}

As real collaborative datasets acquired in savanna are not yet available, for testing the collaborative UAV performance in these wild environments, we simulated multi-agent data input stream starting from some videos recorded in the same area. We used two datasets (Table \ref{table:used_data}), each representing one multi-UAV scenario, collected using a quadrotor DJI Mavic-3E. For dataset 1, image data comes from two flights representing two collaborative aerial agents, whereas in dataset 2 three flights represent three collaborative aerial agents. The flight data selected for each dataset (specifications in Table \ref{table:data_specs}) fulfilled all of the following criteria:
\begin{itemize}
    \item The flights were carried out sequentially or with no more than a 10-minute interval between the end of one flight and start of the next;
    \item the flight trajectories of individual flights were overlapping to test the performance accuracy in merging sparse map of independent flights; 
    \item scenes with moving animals, cars, or people were not used, to focus tests on mapping accuracy .
\end{itemize}

\begin{table}[H]
\begin{tabular}{| P{1.1cm} | P{1.1cm}| P{1.5cm} | P{1.5cm} |P{1.1cm} |} 
  \hline \textbf{Dataset ID} & \textbf{No of agents} & \textbf{V-SLAM frame rate} & \textbf{SfM frame rate} & \textbf{Image size (pixels)} \\ 
  \hline
  1 & 2 & 29.9 fps & 1 fps & 1920 x 1080\\ 
  \hline
  2 & 3 & 29.9 fps & 1 fps & 1920 x 1080\\ 
  \hline
\end{tabular}
\caption{Dataset specifications. For SLAM evaluation we pre-created agent ROSBAG files with messages recorded at 29.97 fps which were played simultaneously on the client nodes to simulate real-time video streaming.} 
\label{table:data_specs}
\end{table}

\noindent In Table \ref{table:dataset_specs} the number of frames used in each processing approach and the flight direction are reported.


\begin{table}[H]
\setlength\tabcolsep{0pt}
\caption{Dataset 1 characteristics: simultaneous flights are simulated from two agents performing data acquisition from opposite viewpoints. Dataset 2 characteristics: three agents were simulated, all flying in the same region, one with different direction with respect to the other two.}
\label{table:dataset_specs}
\begin{tblr}{
  width = \linewidth,
  colspec = {Q[152]Q[127]Q[165]Q[165]Q[353]},
  cells = {c},
  cell{2}{1} = {r=2}{},
  cell{4}{1} = {r=3}{},
  vlines,
  hline{1-2,4,7} = {-}{},
  hline{3,5-6} = {2-5}{},
}
\textbf{Dataset} & \textbf{Agent} & \textbf{Frame Count (V-SLAM)} & \textbf{Frame Count (SfM)} & \textbf{Flight Direction} \\
1                & 1              &      1580                &  55                    &          Towards the Hippo Camp                 \\
                 & 2              &       4685               & 162                     &         Away from the Hippo Camp                  \\
2                & 1              &      6854                &              237        &              Away from the Hippo Camp             \\
                 & 2              &         6855             & 237                     &                  Towards the Hippo Camp         \\
                 & 3              &          6853            & 237                     &       Away from the Hippo Camp                    
\end{tblr}

\end{table}

This selection process enabled us to use the collected datasets to simulate a collaborative mission with multiple drones even if data was acquired by a single UAV performing sequential flights.
 
\section{Methodology} \label{section:methodology}
To investigate the potential and limitations of collaborative mapping, the approaches are considered within two main domains: offline (\ref{subsection:offline_method}) and real-time (\ref{subsection:real_time_method}) processing. Here offline processing refers to mapping performed after complete data acquisition i.e. methods that operate once all data (i.e. images) are available. On the other hand, real-time processing refers to methods that perform mapping while data are being acquired. For already acquired data, real-time processing can be evaluated by simulating a real-time data stream through ROS (for V-SLAM) or terminal clients (for OtF-SfM). To compare the performance of the tested mapping approaches, the available GNSS data (with an accuracy at meter-level in single-point positioning) was used to assess the accuracy of the agent trajectories in terms of Root Mean Square Error (RMSE), i.e. the error between the estimated and the observed GNSS positions.
\subsection{Offline processing}\label{subsection:offline_method}
We selected two well-established SfM software products: a commercial — Agisoft Metashape \footnote{\url{https://www.agisoft.com/}} — and an open-source — COLMAP \cite{schonberger2016structure}, widely used for research purposes due to the significant level of control it offers in all stages of the photogrammetric pipeline. It is not fully known what type of local features for tie point extraction are used in Metashape wheareas COLMAP uses a GPU version of RootSIFT \cite{arandjelovic2012three}. In SfM, distinctive and repeatable keypoints are identified in the images, descriptors are assigned to describe the neighbourhood of these points, and candidate corresponding points (tie points) are exhaustively matched across all pairs of images by comparing the similarity of the descriptors. The candidate matches are further refined using epipolar geometry, from which images can be oriented with different strategies. The incremental approach is considered in this work: after initialization with an image pair, subsequent images are oriented through a resection on the triangulated tie points and expansion of the 3D tie points via triangulation, followed by several local and/or global bundle adjustments. 

Considering that reliable and accurate tie points are fundamental for a good orientation of an image block, keypoints extracted and matched with CNNs have also been tested in addition to the traditional local features implemented in Metashape and COLMAP. These CNN-based features were trained to overcome the limitations of SIFT-like methods, particularly in scenarios involving matching images taken at significantly different viewing angles and with drastic changes in lighting conditions. In the context of this study, the primary challenge is matching images captured by drones following approximately the same trajectory, but in opposite directions, where the terrain is viewed at an angle that varies by approximately 180 degrees. Therefore, SuperPoint\cite{detone_superpoint_2018}, an end-to-end detector and descriptor initially trained on synthetic images with a further refinement on real images, was also included in our tests. SuperPoint features are typically paired with LightGlue\cite{lindenberger_lightglue_2023} or SuperGlue\cite{sarlin_superglue_2020}: we opted for the former due to its faster processing time and more permissive licensing. For these methods, we used the implementation available in the deep-image-matching library\footnote{\url{https://github.com/3DOM-FBK/deep-image-matching/}} (DIM) \cite{morelli2024deep}, which prepares image matching results directly for the import into COLMAP.

\subsection{Real-time processing}\label{subsection:real_time_method}
For real-time collaborative mapping, two distinct approaches were tested. The first is a more photogrammetric approach, referred to as OtF-SfM \cite{zhan_--fly_2024}, while the second is based on V-SLAM, specifically CCM-SLAM \cite{schmuck_ccm-slam_2019}. The OtF-SfM approach utilizes a server-client system, where clients continuously acquire images and send them to a central server. The server groups the images based on similarity and attempts to orient them into a single map. When not feasible, sub-maps are generated and later merged once a sufficient number of common images are available. The incremental mapping process is based on COLMAP, a new fast image retrieval strategy, and a weighted BA based on image similarity.

For the V-SLAM based approach, we chose CCM-SLAM because it is a state-of-the-art collaborative SLAM system that relies only on visual odometry (VO) without inertial sensor fusion. Every VO agent in CCM-SLAM is built on an ORB-SLAM front-end, currently a widely used monocular SLAM solution that limits the amount of observations using only the richest frames through a keyframe selection procedure. Keyframe poses, keypoints, descriptors, and 3D tie points are sent to the server that builds local maps, attempts to close loops, and merges these maps using place recognition\cite{galvez-lopez_bags_2012}  with iterative global bandle adjustments (GBA). The agents download updated keyframe poses from the server after GBA, improving local map accuracy and trajectory estimates with information from other agents.

Since neither OtF-SfM nor CCM currently incorporate GNSS data in trajectory estimation, all results presented here are based solely on image processing. Except for CCM, video frames were selected based on a temporal interval of 1Hz, ensuring approximately 80\% overlap between consecutive images. In the case of CCM, however, the frames were recorded to a ROSBAG file at the frame rate of the video itself which was 29.97 Hz (refer to Table \ref{table:data_specs}). Real-time processing evaluation was performed on an Intel(R) Core(TM) i7-10750H CPU @ 2.60GHz with an Nvidia 2060 Rtx GPU.

\section{Results} \label{section:res_future}

\begin{figure*}[t]
\centering
\begin{tabularx}{\textwidth}{X X}
  \begin{center}
      \includegraphics[width=0.4\textwidth]{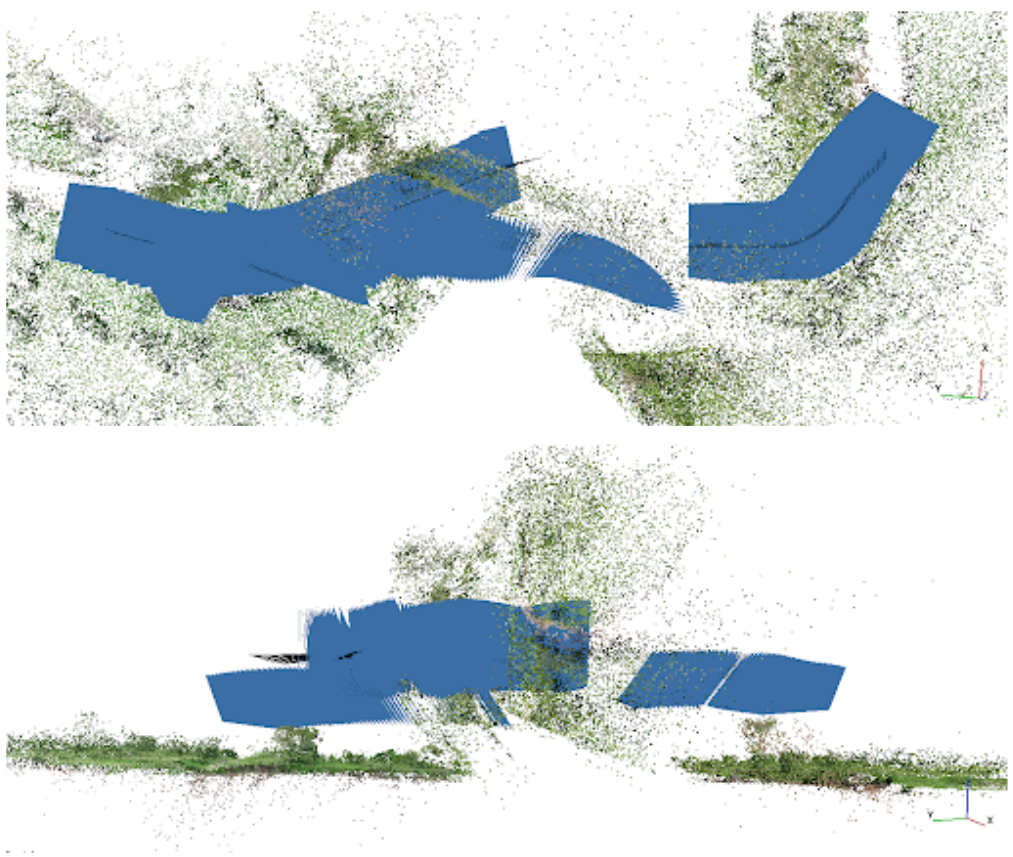} \\
  \end{center}
 &     
 \begin{center}
      \includegraphics[width =0.37\textwidth]{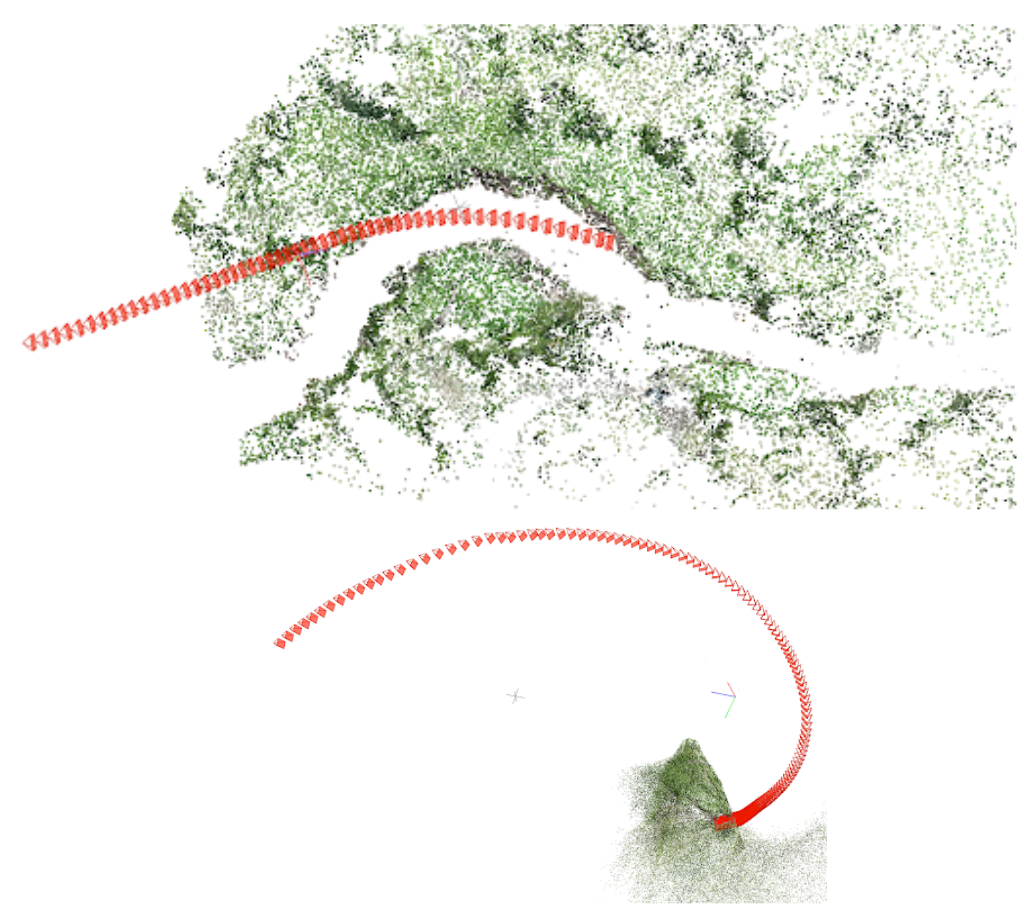}    
 \end{center}
 \\
 
  \multicolumn{1}{c}{(a) Metashape}
      
   & 
  \multicolumn{1}{c}{(b) COLMAP}
  \\
  \begin{center}
      \includegraphics[width=0.43\textwidth]{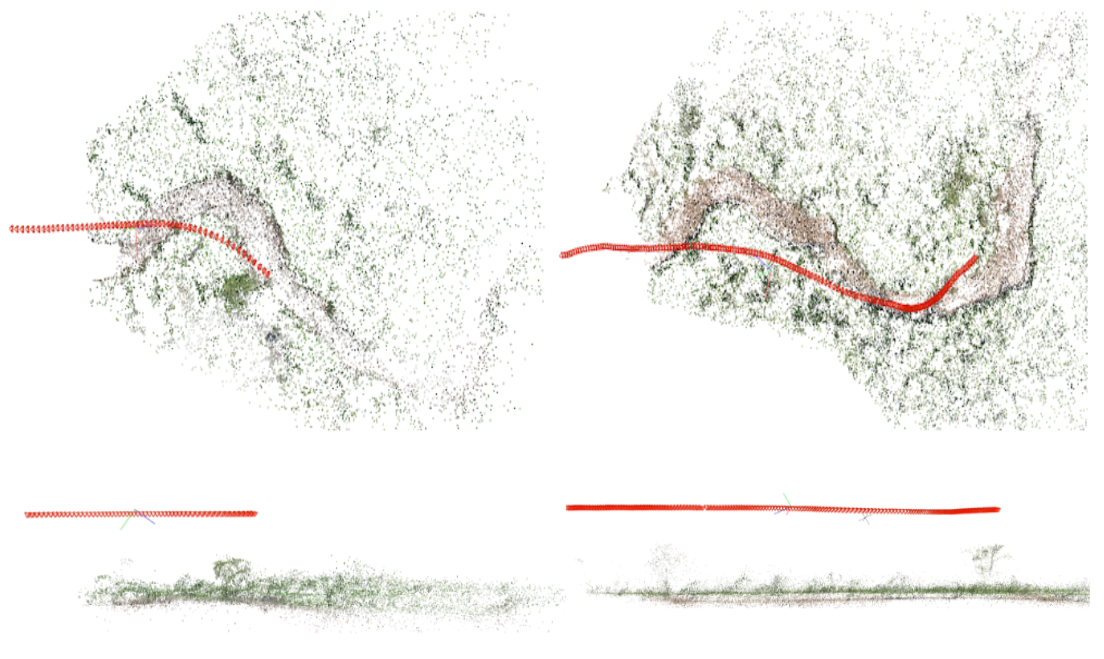} \\
  \end{center}
 &     
 \begin{center}
      \includegraphics[width=0.37\textwidth]{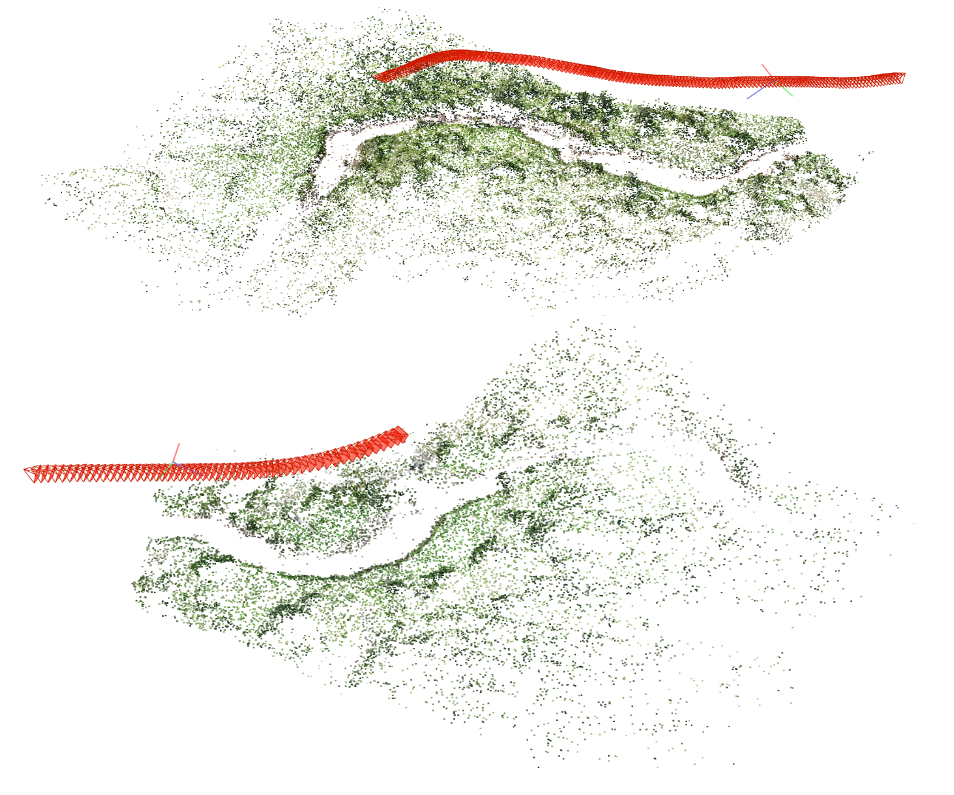}    
 \end{center}
 \\
 \multicolumn{1}{c}{(c) COLMAP + Superpoint}

 &    
 \multicolumn{1}{c}{(d) OtF-SfM} 
 \\
 \begin{center}
      \includegraphics[width=0.38\textwidth]{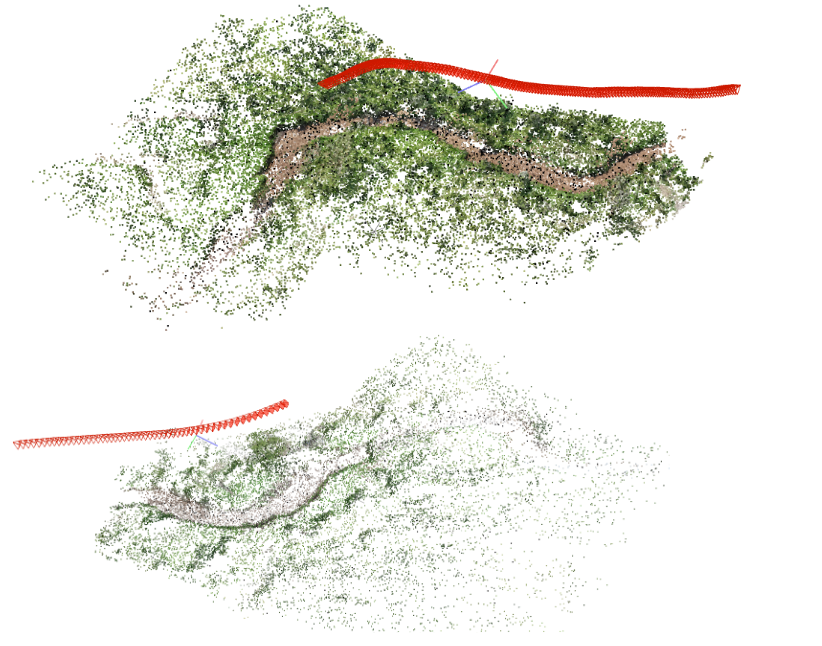} \\
  \end{center}
 &     
 \begin{center}
      \includegraphics[width=0.45\textwidth]{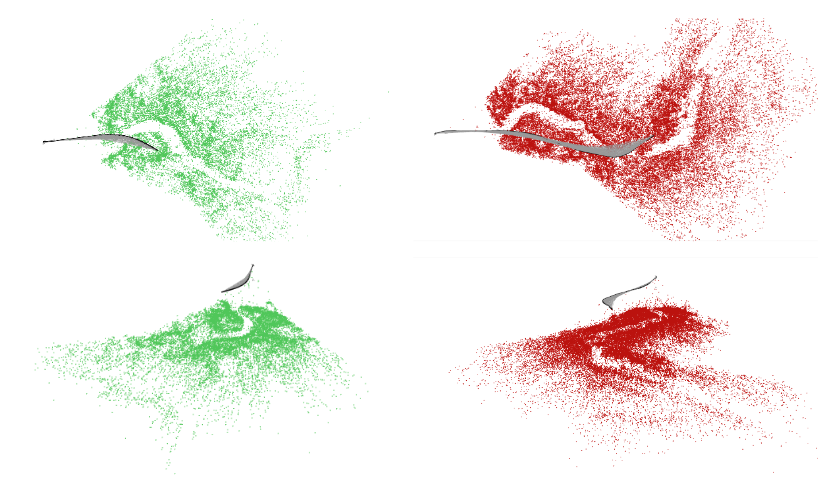}    
 \end{center}
 \\
 \multicolumn{1}{c}{(e) OtF-SFM + SuperPoint}

 &    
 \multicolumn{1}{c}{(f) CCM-SLAM} 
 \\


\end{tabularx}
\caption{Visualization of image orientation results for dataset 1. (a) Degenerate trajectories from Metashape; (b) Degenerate trajectory recovered for agent 1 in COLMAP, shown from top and side view; (c) Trajectories in in COLMAP (SuperPoint) for both agents, shown fr top  and side view; (d) OtF-SFM trajectories for both agents shown from an oblique view point; (e) OtF-SfM (SuperPoint) trajectories for both agents shown from oblique view point; (f) CCM-SLAM trajectories for individual agents shown from top and oblique front view (green point cloud represents mapping through agent 1 and red point cloud from agent 2).}
\label{fig:res_dataset1}
\end{figure*}

\begin{figure*}[t]
\centering
\begin{tabularx}{\textwidth}{X X}
  \begin{center}
      \includegraphics[width=0.38\textwidth]{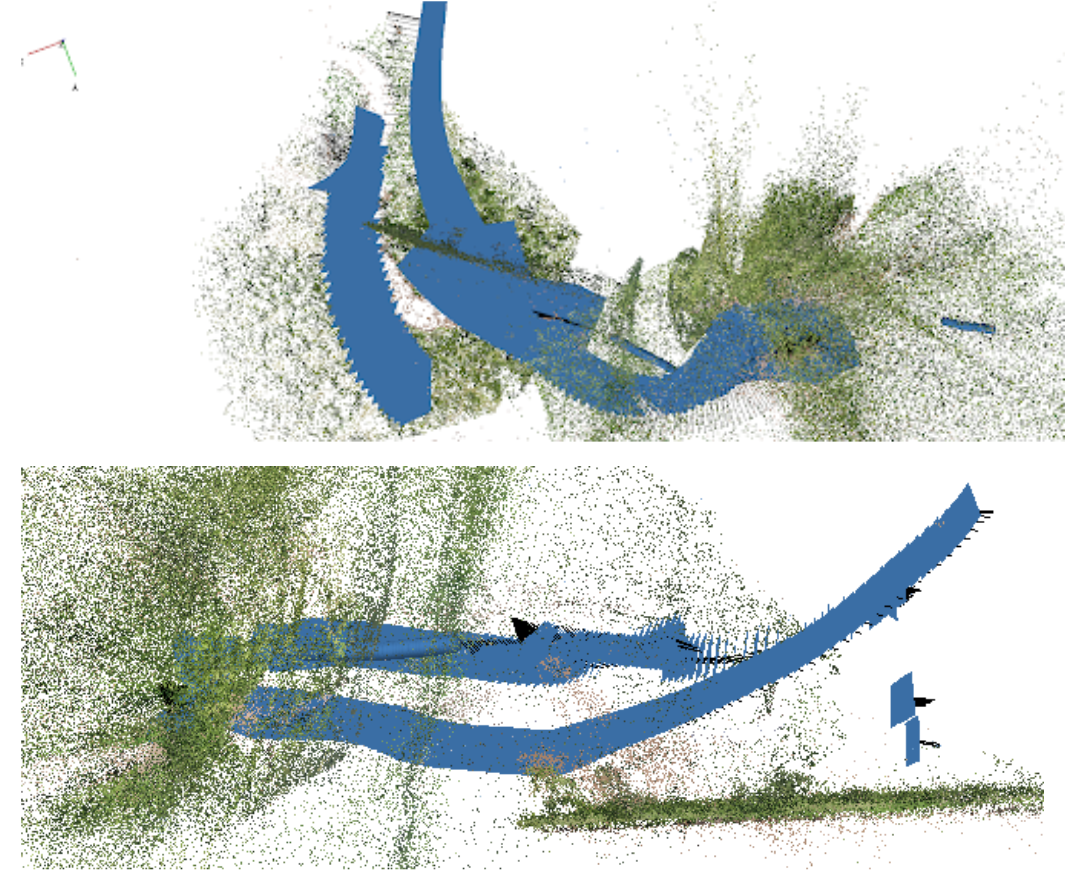} \\
  \end{center}

 &

  \begin{center}
      \includegraphics[width=0.4\textwidth]{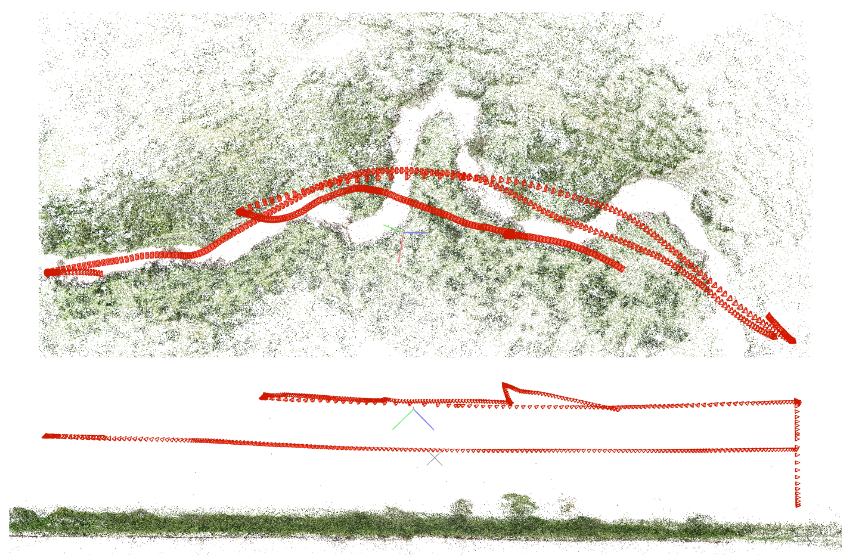} \\
  \end{center}
 
 \\
 
  \multicolumn{1}{c}{(a) Metashape}
      
   & 
  \multicolumn{1}{c}{(b) COLMAP}
  \\
  \begin{center}
      \includegraphics[width=0.37\textwidth]{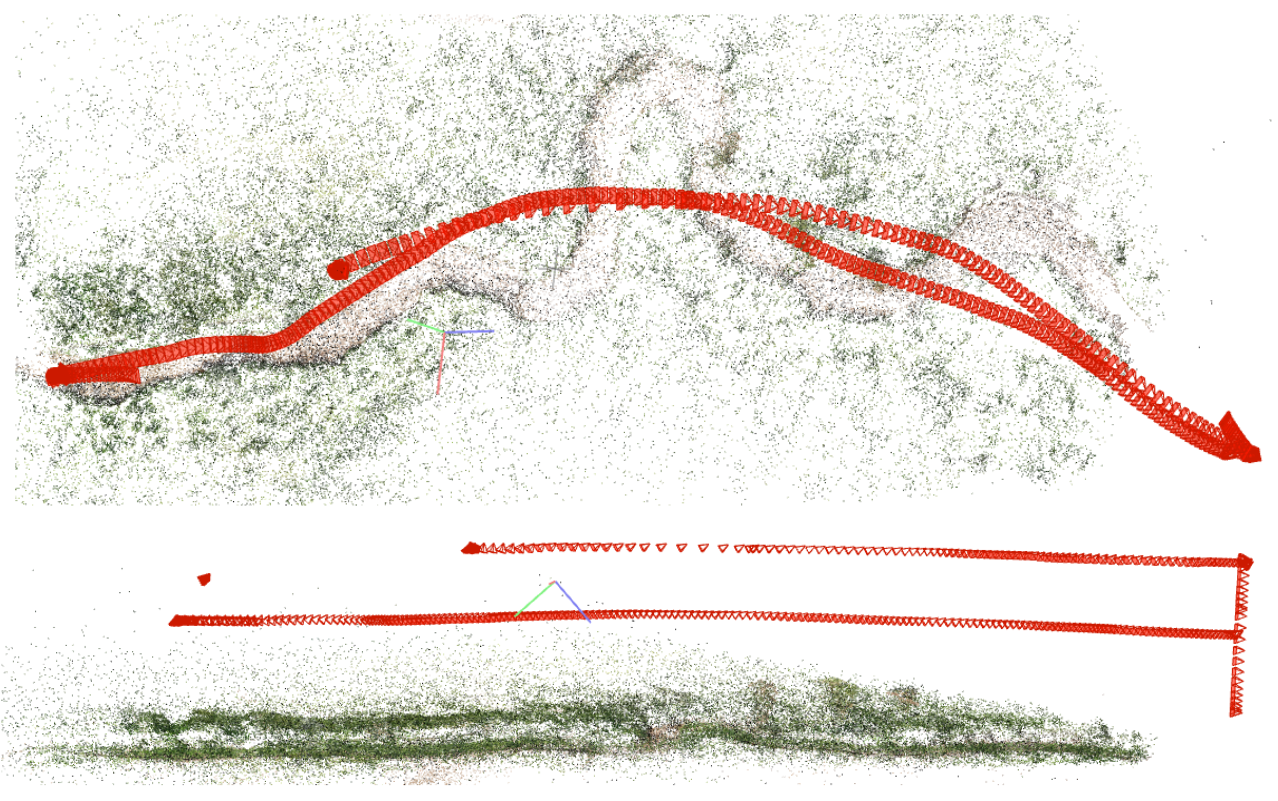} \\
  \end{center}

 &     
 \begin{center}
      \includegraphics[width =0.37\textwidth]{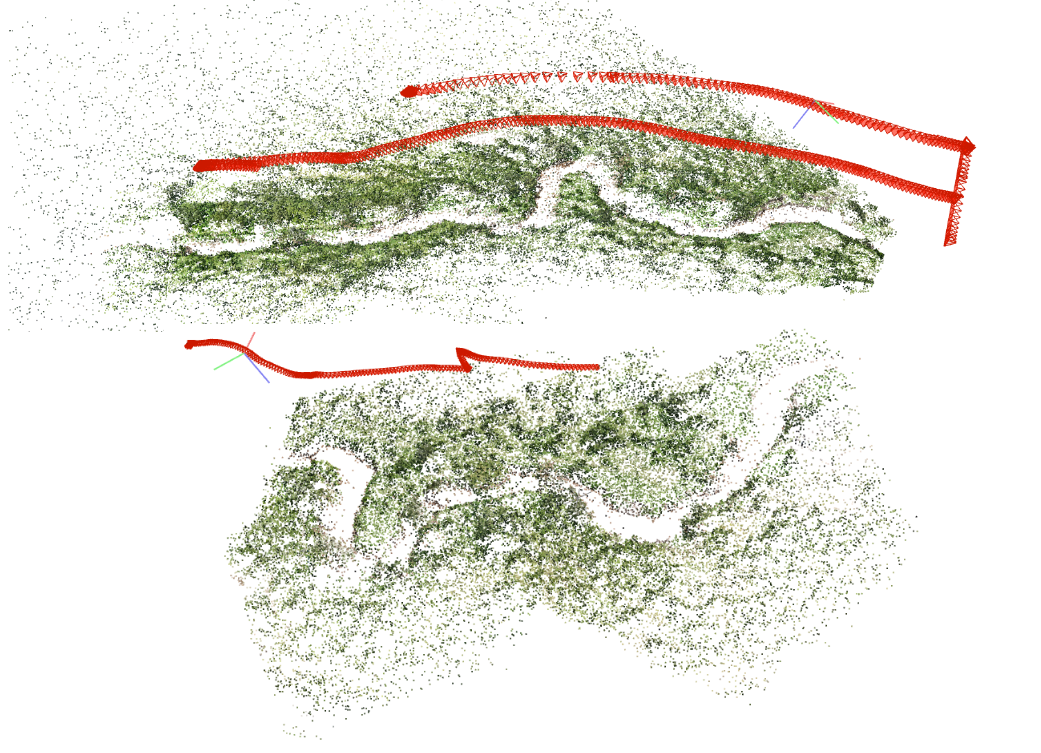}
 \end{center} \\
    \multicolumn{1}{c}{(c) COLMAP + Superpoint}
      &    \multicolumn{1}{c}{(d) OtF-SfM} \\
 \begin{center}
      \includegraphics[width=0.42\textwidth]{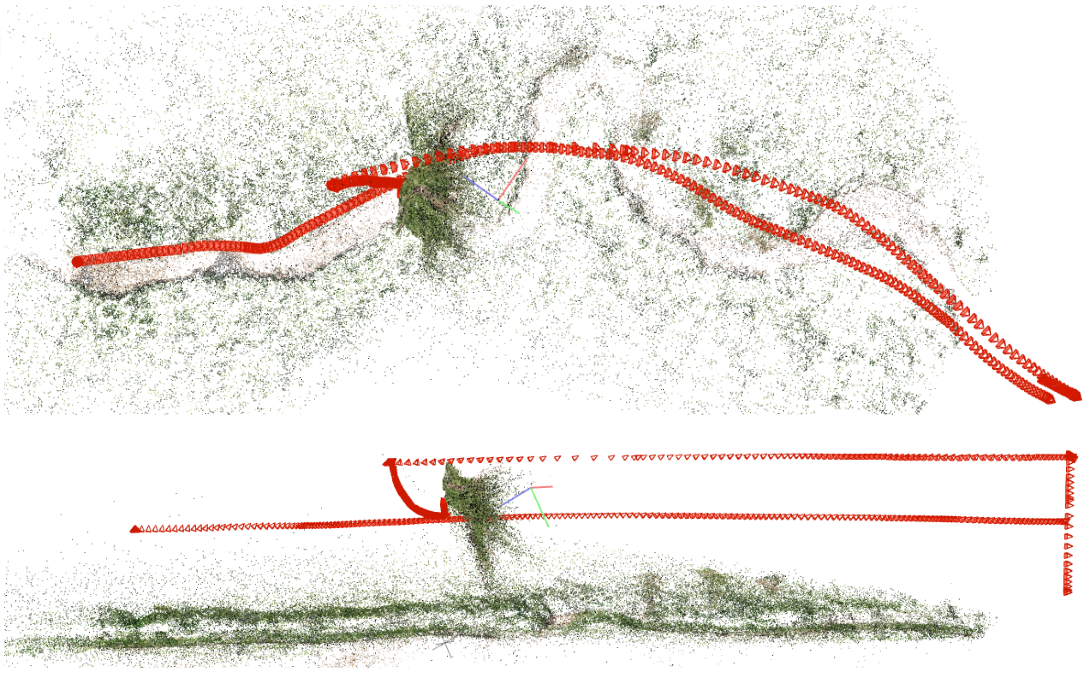} \\
  \end{center}
 &     
 \begin{center}
      \includegraphics[width=0.4\textwidth]{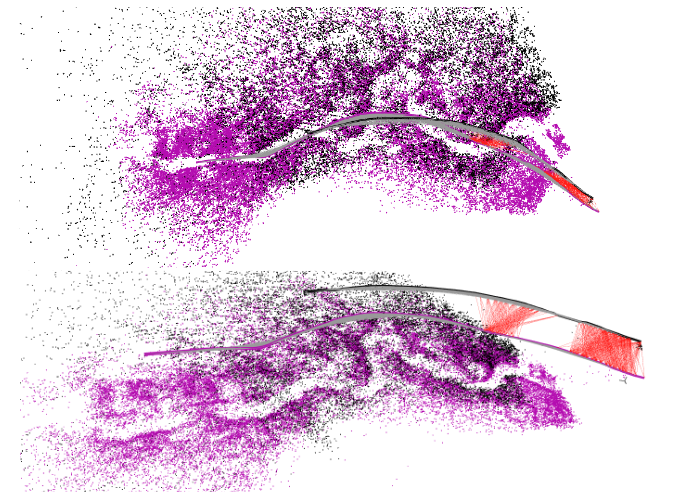}    
 \end{center}
 \\
 \multicolumn{1}{c}{(e) OtF-SfM + SuperPoint}

 &    
 \multicolumn{1}{c}{(f) CCM-SLAM} 
 \\


\end{tabularx}
\caption{Visualization of image orientation results for dataset 2. (a) Degenerate trajectories from Metashape; (b) Trajectories recovered from all three agents in COLMAP shown from nadiral and side view; (c) Trajectories recovered with in COLMAP (SuperPoint) for two agents shown from top and side view; (d) OtF-SfM trajectories for two agents shown from oblique view points; (e) OtF-SfM (SuperPoint) trajectories for two agents shown from nadiral and side view point; (f) Collaborative mapping results in CCM-SLAM for two agents (1 and 3) shown from top and oblique side view (red lines indicate the location matches in key frames).}
\label{fig:res_dataset2}
\end{figure*}

This section presents the processing results for datasets 1 and 2, comparing the estimated trajectories with the GNSS data used as ground truth. RMSE is reported for each agent individually and for the entire dataset when more than one agent trajectories have been estimated together (Tables \ref{table:dataset1_rmse} and \ref{table:dataset2_rmse}).
The processing with traditional SfM software (Metashape and COLMAP used as reference) highlights the complexity of the analyzed datasets, despite their apparent simplicity. At a semantic level, many savanna characteristic elements are well identifiable, such as the river, trees and bushes. At a more granular level, however, the keypoints are quite poor, since they are extracted on the scale of grass and leaves. In both datasets, Metashape failed to correctly orient the image block, creating degenerate trajectories and maps. This happened probably due to the poor quality of the tie points (Figure \ref{fig:res_dataset1}a and \ref{fig:res_dataset2}a). Also COLMAP with RootSIFT features (default) shows degenerate results for dataset 1 (Figure \ref{fig:res_dataset1}b), while for dataset 2 (Figure \ref{fig:res_dataset2}b) it is the only approach that oriented images from all three agents together with the RMSE of 1.10 m (see last column of Table \ref{table:dataset2_rmse}). In general, for all approaches, the major challenge was to orient images from drones flying from the opposite viewpoints. For this reason, COLMAP was tested with the use of SuperPoint, a local feature trained for these kinds of difficult viewing angles. In dataset 1, it succeeds in orienting the agents individually, obtaining the best RMSE (0.20 and 0.11 m), but in dataset 2 it only succeeds in orienting the two agents flying in the same direction and not the third, while COLMAP with RootSIFT succeeded. To understand the possibility of working with variation in flight heights to cater to multiple fieldwork requirements, the results seen so far for SfM processing are all self-calibrated.


\begin{table}[H]
{%
\begin{tabularx}{\linewidth}{|*{3}{C|} }

\hline
\multirow{2}{*}{\textbf{Dataset 1}} & \multicolumn{2}{c|}{\textbf{RMSE [m] on trajectory}}                                                             \\ \cline{2-3} 
                                  & Flight 1 &Flight 2 \\ \hline
\raggedright{Agisoft Metashape}                 & degen    &degen    \\ \hline
\raggedright{COLMAP}                            & degen    &degen    \\ \hline
\raggedright{COLMAP (SuperPoint)}                        & 0.20     &0.11     \\ \hline
\raggedright{OtF-SfM}                        & 0.42     &0.50     \\ \hline
\raggedright{OtF-SfM (SuperPoint)}           & 0.45     &1.18     \\ \hline
\raggedright{CCM-SLAM}                          & 0.30     &0.26     \\ \hline
\end{tabularx}
}
\caption{Dataset 1 results in terms of RMSE of trajectories compared with GNSS positions. Due to critical difference in viewpoint of agent cameras, collaborative mission simulation results were degenerate in all applied methods, hence only single agent trajectories were compared. } 
\label{table:dataset1_rmse}
\end{table}

Collaborative approaches did not produce degenerate re- 
\onecolumn
\setlength\tabcolsep{0pt}

\twocolumn

\begin{table}
\centering
\caption{Dataset 2 results in terms of trajectories RMSEs. Last column describes the number of agents for which trajectories were recovered during collaborative mission simulation.}
\label{table:dataset2_rmse}
\resizebox{\linewidth}{!}{%
\begin{tblr}{
  column{2} = {c},
  column{3} = {c},
  column{4} = {c},
  column{5} = {c},
  cell{1}{1} = {r=2}{},
  cell{1}{2} = {c=5}{},
  vlines,
  hline{1,3-9} = {-}{},
  hline{2} = {2-6}{},
}
\textbf{Dataset 2}      & \textbf{RMSE [m] on Trajectory} &       &       &      &     \\
                        & Fl 1                            & Fl 2  & Fl 3  & Coll & \#  \\
Agisoft Metashape       & degen                           & degen & degen & -    & 0/3 \\
COLMAP                  & 0.39                            & 0.27  & 0.30  & 1.10 & 3/3 \\
COLMAP (SuperPoint)              & 0.28                            & 0.27  & 0.36  & 1.03 & 2/3 \\
On-the-Fly              & 1.52                            & 0.80  & 0.98  & 1.54 & 2/3 \\
On-the-Fly (SuperPoint) & degen                           & 1.51  & 0.69  & 1.46 & 2/3 \\
CCM-SLAM                & 0.74                            & 0.54  & 0.81  & 6.39 & 2/3 
\end{tblr}
}
\end{table}

\begin{figure}[hbt]
\centering
\includegraphics[width=1.0\columnwidth]{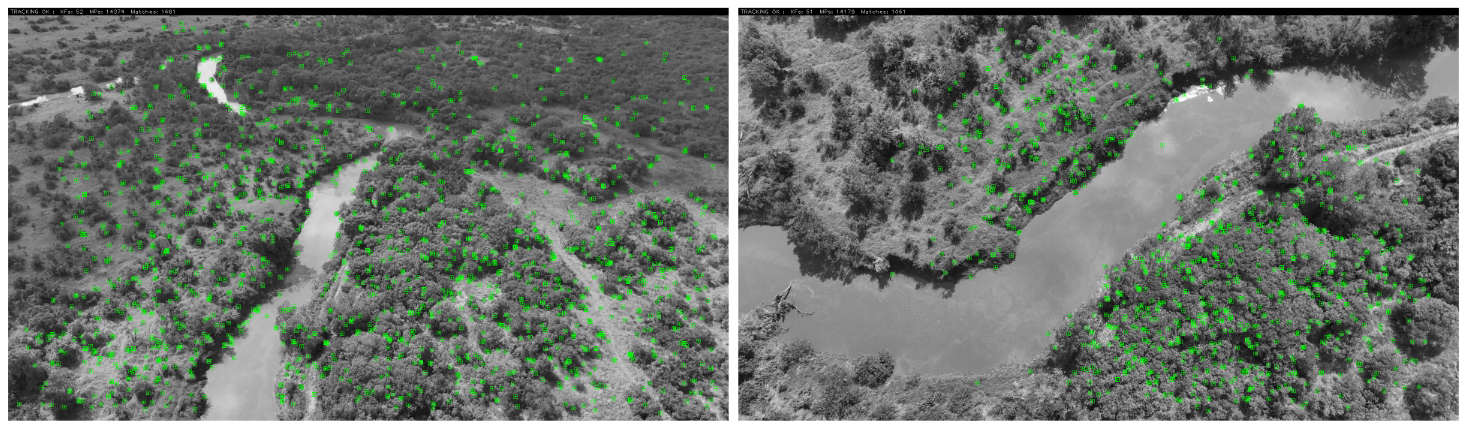}
    \caption{CCM-SLAM loses tracking of features when the camera undergoes rotation movement. RQt visualization shows the reduction in tracked features as the drone turns.}
    \label{fig:eval}
\end{figure}

 \noindent constructions, probably thanks to mandating the use of pre-calibrated cameras, which reduced the influence of outliers and inaccuracies in tie points as they were easily filtered out. Furthermore, contrary to traditional SfM methods, they generated trajectories and maps in real-time, but generally at the cost of higher RMSE (Table \ref{table:dataset1_rmse} and \ref{table:dataset2_rmse}). Neither OtF-SfM with SIFT or SuperPoint, nor CCM-SLAM managed to orient agents flying in opposite directions due to the strong change in viewing angle. This problem seems to be related to the performance of the local features and matchers used and the specific implementation, while COLMAP SfM with RootSIFT had been able to orient the entire block. In terms of accuracy, CCM-SLAM was more accurate in dataset 1 than OtF, in particular in the case of using SuperPoint. But in dataset 2, OtF achieved much better accuracy with respect to CCM-SLAM, and with a more complete trajectory. In fact, in CCM-SLAM, upon a critical rotation of agent, tracked features were lost partially (Figure \ref{fig:eval}) or entirely. This was observed mostly upon fast rotation of camera, swift change in camera pitch or due to small-parallax motion in case of high-altitude UAV flight, such as is performed in wildlife habitat surveys. Since CCM-SLAM cannot recover from lost tracking, parts of the trajectory were lost. 

\section{Conclusions} \label{section:conclus}
The paper presented results and challenges of UAV-based collaborative 3D mapping for monitoring large-scale savanna environments. The results show that relying solely on visual data is promising, with open-source collaborative frameworks available for real-time processing. However, the solution is prone to the quality of local feature matches. To solve these issues, the addition of IMU or RTK-enabled GNSS data could be considered to avoid mapping errors and degenerate solutions.

When compared to visual-inertial odometry (VIO), purely video based visual odometry (VO) is less common in applications of V-SLAM in outdoor environments \cite{danping_collaborative_2019}. VO can be sensitive to repetitive textures, dynamic scenes and is more prone to tracking losses or initialization failures. Inertial measurements make outdoor operations more robust to challenging situations, compared to VO \cite{cremona_experimental_2022}.

However, ensuring that a comparable quality of results is achieved with just visual data, a purely photogrammetric approach has greater potential for widespread use in widelife research. This is due to the popularity and high availability of off-the-shelf drones, which can easily provide high-quality video sequences, but inertial measurements and RTK GNSS positions can be hard or impossible to retrieve from their data logs. Therefore, evaluating the performance of real-time camera-based pipelines such as Visual-SLAM or OtF-SfM is crucial to understand their applicability to UAV swarm technology for wildlife conservation through habitat monitoring. 

\section*{Acknowledgements}
The WildDrone project \url{https://wilddrone.eu/} has received funding from the European Union’s Horizon Europe research and innovation programme under the Marie Skłodowska-Curie grant agreement no. 101071224.


\bibliographystyle{unsrt}
\bibliography{full_paper}

\begin{thebibliography}{10}

\bibitem{tuia_perspectives_2022}
D.~Tuia, B.~Kellenberger, S.~Beery, B.R. Costelloe, S.~Zuffi, B.~Risse, A.~Mathis, M.W. Mathis, F.~van Langevelde, T.~Burghardt, R.~Kays, H.~Klinck, M.~Wikelski, I.D. Couzin, G.~van Horn, M.~C. Crofoot, C.V. Stewart, and T.~Berger-Wolf.
\newblock Perspectives in machine learning for wildlife conservation.
\newblock {\em Nature Communications}, 13(1):792.

\bibitem{wirsing_2022}
A.~Wirsing, A.~Johnston, and J.~Kiszka.
\newblock The rapidly expanding role of drones as a tool for wildlife research.
\newblock {\em Wildlife Research}, 49, 02 2022.

\bibitem{shapiro_conservation_2020}
A.~Shapiro, K.~Anderson, J.~Duffy, F.~Spina~Avino, L.~{DeBell}, and P.~Glover-Kapfer.
\newblock {\em Conservation Technology Series Issue 5: {DRONES} {FOR} {CONSERVATION}}.

\bibitem{shukla_towards_2024}
V.~Shukla, L.~Morelli, F.~Remondino, A.~Micheli, D.~Tuia, and B.~Risse.
\newblock Towards estimation of 3d poses and shapes of animals from oblique drone imagery.
\newblock {\em Int. Arch. Photogramm. Remote Sens. Spatial Inf. Sci.}, {XLVIII}-2-2024:379--386.

\bibitem{barbeau_research_2022}
M.~Barbeau, J.~Garcia-Alfaro, and E.~Kranakis.
\newblock Research trends in collaborative drones.
\newblock {\em Sensors}, 22(9):3321.

\bibitem{10318509}
H.~Wang, J.~Tang, Q.~Pan, Z.~Zhao, and Z.~Wang.
\newblock Multi-uav collaborative reconnaissance based on adaptive particle swarm optimization.
\newblock In {\em 2023 IEEE Proc (ICUS)}, pages 559--564, 2023.

\bibitem{huang_multi-uav_2022}
R.~Huang, H.~Zhou, T.~Liu, and H.~Sheng.
\newblock Multi-{UAV} collaboration to survey tibetan antelopes in hoh xil.
\newblock {\em Drones}, 6(8):196.

\bibitem{kabir_wildlife_2021}
R.H. Kabir and K.~Lee.
\newblock Wildlife monitoring using a multi-{UAV} system with optimal transport theory.
\newblock {\em Applied Sciences}, 11(9):4070.

\bibitem{tong_multi-uav_2023}
P.~Tong, X.~Yang, Y.~Yang, W.~Liu, and P.~Wu.
\newblock Multi-{UAV} collaborative absolute vision positioning and navigation: A survey and discussion.
\newblock {\em Drones}, 7(4):261.

\bibitem{chen_overview_2023}
W.~Chen, X.~Wang, S.~Gao, G.~Shang, C.~Zhou, Z.~Li, C.~Xu, and K.~Hu.
\newblock Overview of multi-robot collaborative {SLAM} from the perspective of data fusion.
\newblock {\em Machines}, 11:653.

\bibitem{10.1145/3150165.3150166}
F.~Poiesi, A.~Locher, P.~Chippendale, E.~Nocerino, F.~Remondino, and L.~Van~Gool.
\newblock Cloud-based collaborative 3d reconstruction using smartphones.
\newblock In {\em Proc CVMP}, CVMP '17, 2017.

\bibitem{quantifying}
B.~Koger, A.~Deshpande, J.T. Kerby, J.M. Graving, B.R. Costelloe, and I.D. Couzin.
\newblock Quantifying the movement, behaviour and environmental context of group-living animals using drones and computer vision.
\newblock {\em Journal of Animal Ecology}, 92(7):1357--1371, 2023.

\bibitem{wang_uav-based_2024}
K.~Wang, L.~Kooistra, R.~Pan, W.~Wang, and J.~Valente.
\newblock {UAV}-based simultaneous localization and mapping in outdoor environments: A systematic scoping review.
\newblock {\em Journal of Field Robotics}.

\bibitem{davison_monoslam_2007}
A.J. Davison, I.D. Reid, N.D. Molton, and O.~Stasse.
\newblock {MonoSLAM}: Real-time single camera {SLAM}.
\newblock {\em {IEEE} Transactions on Pattern Analysis and Machine Intelligence}, 29(6):1052--1067.

\bibitem{ahmed_active_2023}
M.F. Ahmed, K.~Masood, V.~Fremont, and I.~Fantoni.
\newblock Active {SLAM}: A review on last decade.
\newblock {\em Sensors}, 23(19):8097.

\bibitem{cao_scaling_2024}
H.~Cao, J.~Xu, Z.~Yang, L.~Shangguan, J.~Zhang, X.~He, and Y.~Liu.
\newblock Scaling up edge-assisted real-time collaborative visual {SLAM} applications.
\newblock {\em {IEEE}/{ACM} Transactions on Networking}, 32(2):1823--1838.

\bibitem{danping_collaborative_2019}
Z.~Danping, P.~Tan, and W.~Yu.
\newblock Collaborative visual {SLAM} for multiple agents:a brief survey.
\newblock {\em Virtual Reality \& Intelligent Hardware}, 1:461--482.

\bibitem{jiang_efficient_2020}
S.~Jiang, C.~Jiang, and W.~Jiang.
\newblock Efficient structure from motion for large-scale {UAV} images: A review and a comparison of {SfM} tools.
\newblock {\em {ISPRS} Journal of Photogrammetry and Remote Sensing}, 167:230--251.

\bibitem{reja_computer_2022}
V.K. Reja, K.~Varghese, and Q.P. Ha.
\newblock Computer vision-based construction progress monitoring.
\newblock {\em Automation in Construction}, 138:104245.

\bibitem{zhan_--fly_2024}
Z.~Zhan, R.~Xia, Y.~Yu, Y.~Xu, and X.~Wang.
\newblock On-the-fly {SfM}: What you capture is what you get.
\newblock {\em {ISPRS} Annals of the Photogrammetry, Remote Sensing and Spatial Information Sciences}, X-1-2024:297--304.

\bibitem{9812319}
Y.~Yue, C.~Zhao, Y.~Wang, Y.~Yang, and D.~Wang.
\newblock Aerial-ground robots collaborative 3d mapping in gnss-denied environments.
\newblock In {\em Proc. ICRA}, pages 10041--10047, 2022.

\bibitem{schmuck_ccm-slam_2019}
P.~Schmuck and M.~Chli.
\newblock {CCM}-{SLAM}: Robust and efficient centralized collaborative monocular simultaneous localization and mapping for robotic teams.
\newblock {\em Journal of Field Robotics}, 36(4):763--781.

\bibitem{morelli2024deep}
L.~Morelli, F.~Ioli, F.~Maiwald, G.~Mazzacca, F.~Menna, and F.~Remondino.
\newblock Deep-image-matching: a toolbox for multiview image matching of complex scenarios.
\newblock {\em Int. Arch. Photogramm. Remote Sens. Spatial Inf. Sci.}, 48:309--316, 2024.

\bibitem{schonberger2016structure}
J.L. Schonberger and Jan-Michael Frahm.
\newblock Structure-from-motion revisited.
\newblock In {\em Proc CVPR}, pages 4104--4113, 2016.

\bibitem{arandjelovic2012three}
R.~Arandjelovi{\'c} and A.~Zisserman.
\newblock Three things everyone should know to improve object retrieval.
\newblock In {\em 2012 IEEE Proc CVPR}, pages 2911--2918. IEEE, 2012.

\bibitem{detone_superpoint_2018}
D.~{DeTone}, T.~Malisiewicz, and A.~Rabinovich.
\newblock {SuperPoint}: Self-supervised interest point detection and description.
\newblock In {\em Proc. CVPR 2018}, pages 337--33712.

\bibitem{lindenberger_lightglue_2023}
P.~Lindenberger, P.-E. Sarlin, and M.~Pollefeys.
\newblock {LightGlue}: Local feature matching at light speed.
\newblock In {\em Proc. ICCV 2023}, pages 17581--17592.

\bibitem{sarlin_superglue_2020}
Paul-Edouard Sarlin, D.~{DeTone}, T.~Malisiewicz, and A.~Rabinovich.
\newblock {SuperGlue}: Learning feature matching with graph neural networks.
\newblock In {\em Proc. CVPR 2020}, pages 4937--4946.

\bibitem{galvez-lopez_bags_2012}
D.~Galvez-López and J.D. Tardos.
\newblock Bags of binary words for fast place recognition in image sequences.
\newblock {\em {IEEE} Transactions on Robotics}, 28(5):1188--1197.

\bibitem{cremona_experimental_2022}
J.~Cremona, R.~Comelli, and T.~Pire.
\newblock Experimental evaluation of visual-inertial odometry systems for arable farming.
\newblock {\em Journal of Field Robotics}, 39(7):1121--1135.

\end{thebibliography}

\end{document}